\documentclass{article}
\usepackage{spconf,amsmath,graphicx,hyperref}

\usepackage{booktabs}  
\usepackage{array}     
\usepackage[numbers]{natbib}
\usepackage[table]{xcolor}
\usepackage{pgfplots}
\pgfplotsset{compat=1.18}
\usepackage{tikz}
\usepackage{todonotes}
\usepackage{tabularx}

\definecolor{verbatimblue}{RGB}{0,70,150}

\usepackage{arabtex}
\usepackage{utf8}
\setcode{utf8}

\usepackage{multirow}
\usepackage{subcaption}

\DeclareRobustCommand{\aranextgenvoices}{\textsc{AraYoungVoices}}

\DeclareRobustCommand{\aranextgentitle}{%
  {\color{verbatimblue}AraYoungVoices}%
}
\title{\aranextgentitle:
A Diverse L1/L2 Corpus of \\ Arabic Child and Adolescent Speech}
\name{Shammur Absar Chowdhury, Zien Sheikh Ali, Houssam Eddine-Othman Lachemat, Hamdy Mubarak%
\thanks{The work is supported by the HBKU Flagship Research Grant
(HBKU-INT-VPR-FRG-03-09).}}

\address{Qatar Computing Research Institute, Qatar}%
\begin{document}
\ninept
\maketitle
%

\begin{abstract}
State-of-the-art ASR systems primarily target native adult speech, leading to substantial performance gaps for children, adolescents, and L2 speakers. We introduce \textbf{\aranextgenvoices}, a 151.72-hour Arabic read-speech corpus from 286 speakers aged 7--18, comprising \textsc{AraKids} (7--12) and \textsc{AraTeens} (13--18). The corpus includes 146 native Arabic (L1) and 140 second-language (L2) speakers, with native speakers spanning
Egyptian, Gulf, Levantine, and North African dialectal backgrounds and L2 speakers representing diverse linguistic backgrounds across the Americas, Asia, Africa, and Europe. We benchmark four pretrained ASR models under zero-shot and fine-tuned settings using unseen-speaker-$\&$-unseen-prompt
(\textsc{USUP}) and unseen-speaker-$\&$-seen-prompt (\textsc{USSP}) evaluations. Results show that L2 speech remains substantially more challenging than L1 speech, with the largest errors observed mainly for younger L2 speakers. Age-specific fine-tuning improves the matched age group, while joint fine-tuning provides a stronger balance across populations. ASR hypotheses are also consistently closer to the standard reading prompt than to the verbatim transcription, particularly for L2 speech, suggesting partial normalization of reading deviations. 
\end{abstract}


%

\section{Introduction}
\label{sec:intro}

Automatic speech recognition (ASR) has improved significantly with large pretrained multilingual speech models trained on massive amounts of speech data. However, performance remains uneven across age groups and language backgrounds. Children, adolescents, and second-language (L2) speakers remain particularly challenging because pronunciation, fluency, and acoustic characteristics differ from native adult speech and evolve with
age \cite{gerosa2009review,zhou2025childmandarin}. Adolescent speech, which captures the transition from child to adult speech, remains comparatively understudied.

Existing child-speech corpora cover several languages and speaking styles. Early resources such as OGI Kids, PF-STAR, CHOREC, and SingaKids supported research on child ASR
\cite{shobaki00_icslp,batliner05b_interspeech, cleuren-etal-2008-childrens,chen16g_interspeech}, while more recent corpora include MyST, BK3AT, ChildMandarin, and ChildTalk \cite{pradhan-etal-2024-science,gonzales-etal-2024-bk3at,
zhou-etal-2025-childmandarin,zhou-etal-2026-childtalk}.
Resources on adolescence or L2 speech are fewer comparatively
\cite{gretter-etal-2020-tlt,meister2023developmental,draxler06_interspeech}.
For Arabic, available child-speech resources remain limited. Arabic Little STT,
for example, contains only 355 utterances from children aged 6--13
\cite{alkadri2025arabiclittle}. Table~\ref{tab:child-adolescent-corpora} summarize the current resources.

\begin{figure}[!ht]
  \centering
  \includegraphics[width=0.98\columnwidth]{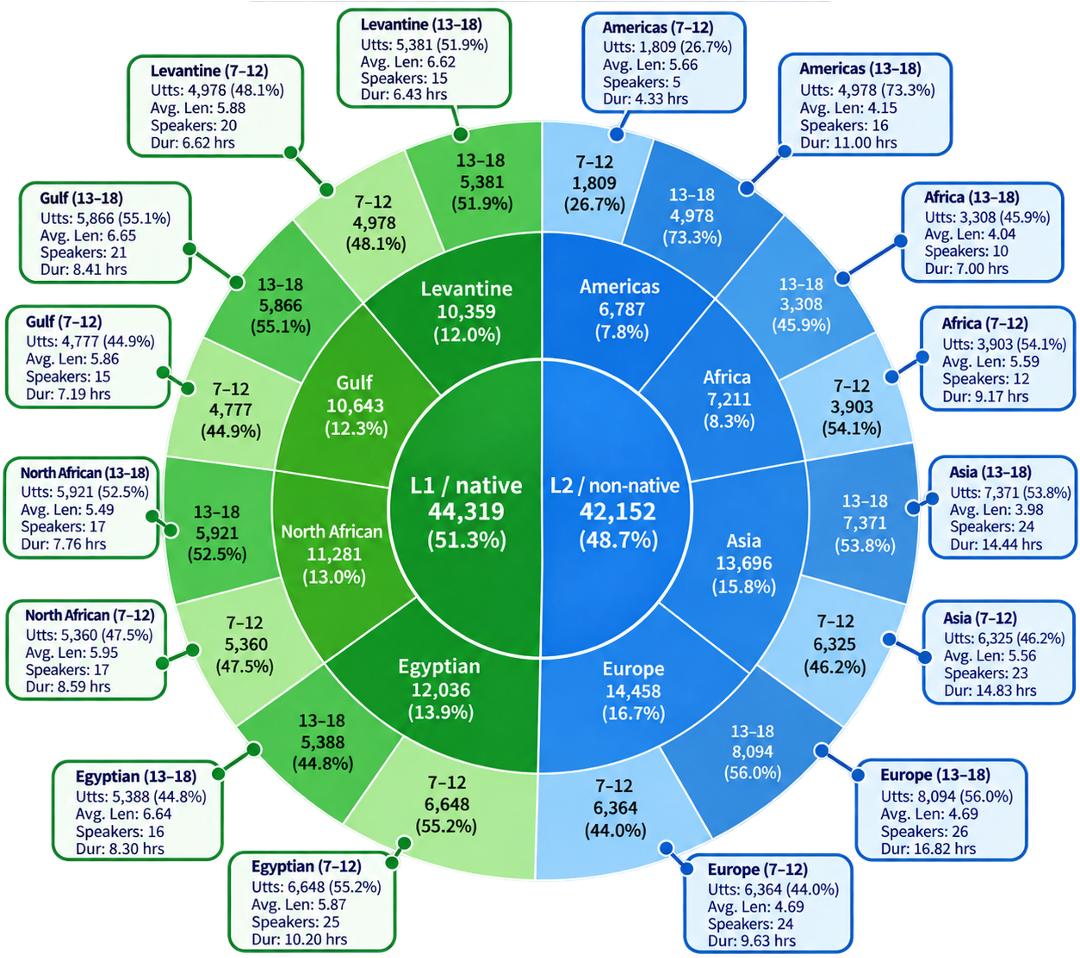}
  \vspace{-0.2cm}
  \caption{Overview of \aranextgenvoices{}. Avg.\ Len.: average prompt length in words. }
  \label{fig:corpus_overview}
  \vspace{-0.2cm}
\end{figure}

\begin{table*}[!ht]
\centering
\caption{Available Child and Adolescent speech corpora grouped by age
coverage. K: kindergarten, G: grade.}
\label{tab:child-adolescent-corpora}

\scriptsize
\setlength{\tabcolsep}{2.5pt}
\renewcommand{\arraystretch}{0.88}

\begin{tabular*}{\textwidth}{
@{\extracolsep{\fill}}
l l c r r
@{\hspace{6pt}}
l l c r r
@{}}

\toprule

\multicolumn{5}{c}{\textbf{Primarily Child-Focused}} &
\multicolumn{5}{c}{\textbf{Child--Adolescent / Adolescent}} \\

\cmidrule(lr){1-5}
\cmidrule(lr){6-10}

\textbf{Corpus} &
\textbf{Language} &
\textbf{Age} &
\textbf{Speakers} &
\textbf{Hours} &
\textbf{Corpus} &
\textbf{Language} &
\textbf{Age} &
\textbf{Speakers} &
\textbf{Hours} \\

\midrule

CHOREC \cite{cleuren-etal-2008-childrens}
& NL & 6--12 & 400 & 130 &
OGI Kids \cite{shobaki00_icslp}
& EN & K--G10 & 1,100 & -- \\

SingaKids \cite{chen16g_interspeech}
& ZH & 7--12 & 255 & 75 &
PF-STAR \cite{batliner05b_interspeech}
& Multi. & 4--14 & -- & $>$60 \\

ChildMandarin \cite{zhou-etal-2025-childmandarin}
& ZH & 3--5 & 397 & 41.25 &
TLT-school \cite{gretter-etal-2020-tlt}
& L2 EN/DE & 9--16 & 2,710 & -- \\

MyST \cite{pradhan-etal-2024-science}
& EN & G3--G5 & $\sim$1,300 & $\sim$400 &
Estonian Adolescent \cite{meister2023developmental}
& ET & 9--18 & 309 & $\sim$70 \\

BK3AT \cite{gonzales-etal-2024-bk3at}
& Multi. & G1--G3 & -- & 122 &
Ph@ttSessionz \cite{draxler06_interspeech}
& DE & 12--20 & $\sim$1,000 & -- \\

ChildTalk$^{a}$ \cite{zhou-etal-2026-childtalk}
& ZH & 2--8 & 498 & 112.5 &
\textcolor{verbatimblue}{\textbf{\aranextgenvoices{}}}
& \textcolor{verbatimblue}{\textbf{AR}} & \textcolor{verbatimblue}{\textbf{7--18}} & \textcolor{verbatimblue}{\textbf{286}} & \textcolor{verbatimblue}{\textbf{151.72}} \\

Arabic Little STT \cite{alkadri2025arabiclittle}
& AR & 6--13 & 288 & -- &
& & & & \\

\bottomrule
\end{tabular*}

\vspace{0.3mm}
\begin{minipage}{0.98\textwidth}
\end{minipage}
\vspace{-0.3cm}
\end{table*}

This gap is particularly important for Arabic, where speech varies across major dialectal backgrounds such as Egyptian, Gulf, Levantine, and North African Arabic, while Modern Standard Arabic is primarily used in formal and
educational settings \cite{ali_connecting_2021,bouamor2018madar}. For L2 speakers, pronunciation is further influenced by first language, proficiency, and other language exposure. Consequently, robust speech models for young Arabic speakers therefore must account for age, L1/L2 background, and dialectal variation.

To reduce the gap, we introduce \aranextgenvoices{}, an Arabic read-speech corpus containing \textbf{151.72 hours}, \textbf{86,471 utterances}, and \textbf{286 speakers} \textbf{aged 7--18.} It comprises \textsc{AraKids} (7--12) and
\textsc{AraTeens} (13--18), with \textbf{146 native Arabic} (L1) and \textbf{140 non-native speakers} (L2).
L1 speakers comes from Egyptian, Gulf, Levantine, and North African dialectal backgrounds, while L2 speakers represent diverse language backgrounds across Africa, the Americas, Asia, and Europe (see Figure~\ref{fig:corpus_overview}).

To understand the effectiveness of the corpus, we evaluate contemporary ASR models in zero-shot and fine-tuned settings using two complementary test conditions: \textsc{USUP}, with unseen speakers and
unseen prompts, and \textsc{USSP}, with unseen speakers reading prompts observed during training. Our analysis examines how recognition varies across age, L1/L2 background, and dialect, and whether age-specific or joint fine-tuning reduces these gaps.

\noindent Our contributions are: \textit{(i)} \textbf{\aranextgenvoices{}}, a
large-scale Arabic corpus spanning children and adolescents across diverse L1
and L2 backgrounds; \textit{(ii)} age- and nativity-aware ASR benchmarks with
complementary speaker and prompt generalization settings; \textit{(iii)} analysis of recognition gaps across age, L1/L2 status, and dialectal backgrounds; and \textit{(iv)} release\footnote{Link removed for anonymity.}
of fine-tuned ASR models for young Arabic speech and along with annotation guidelines.

\begin{figure}[h]
    \centering

    \begin{subfigure}[t]{\columnwidth}
        \centering
        \includegraphics[width=\linewidth]{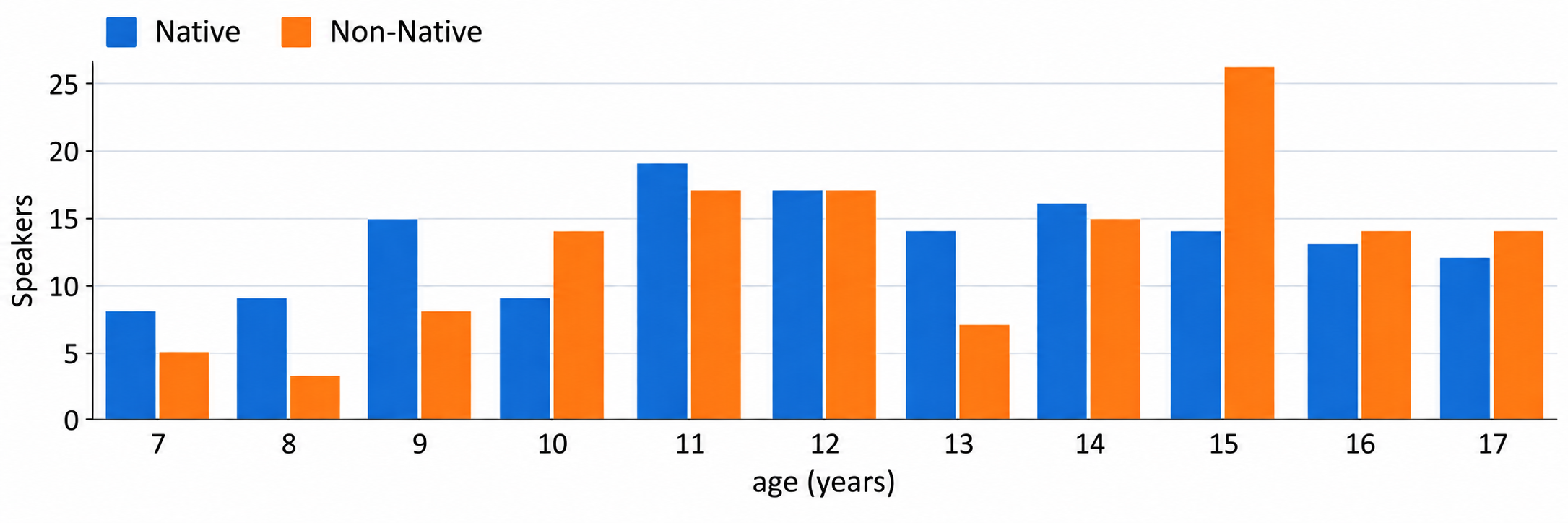}
        \vspace{-0.5cm}
        \caption{Age and nativity.}
    \end{subfigure}

%
    \begin{subfigure}[t]{\columnwidth}
        \centering
        \includegraphics[width=\linewidth]{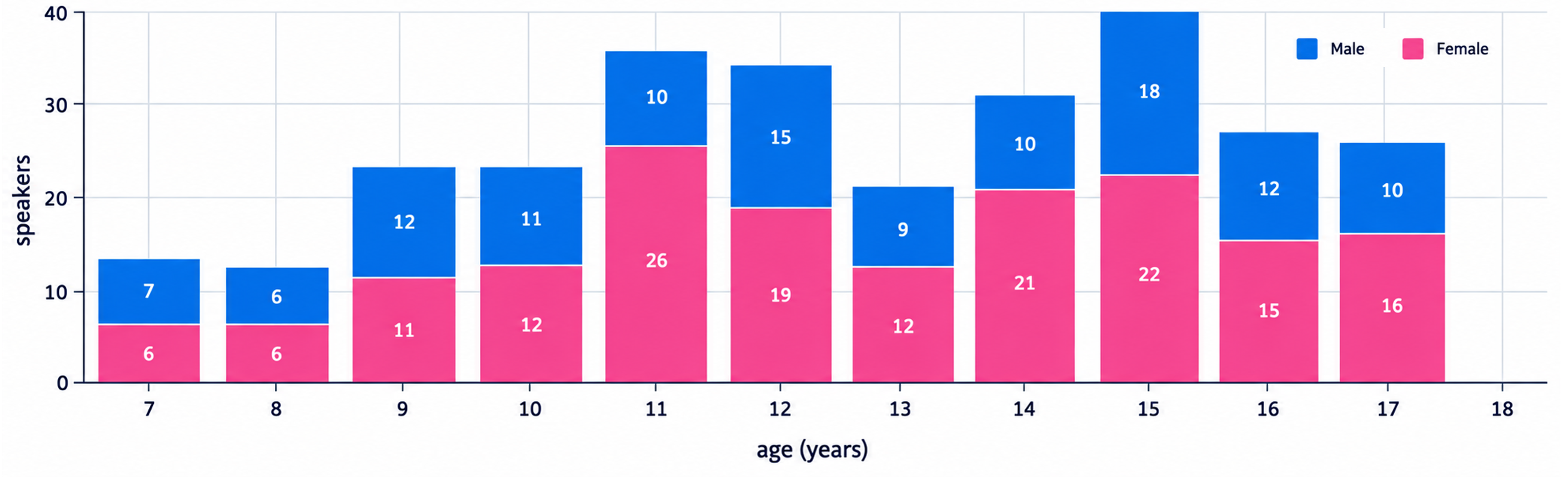}
        \vspace{-0.5cm}
        \caption{Age and gender.}
    \end{subfigure}



    \caption{Speaker  characteristics in \aranextgenvoices{}.}
    \label{fig:dataset_characteristics}
    \vspace{-0.35cm}
\end{figure}

\section{\aranextgenvoices{}}
\label{sec:data}


\aranextgenvoices{} is an Arabic read-speech corpus designed to study variation across age and language background. It contains \textbf{151.72 hours} of speech and 86,471 utterances from \textbf{286 speakers} aged 7--18. The corpus comprises two age groups: \textsc{AraKids} (7--12) and \textsc{AraTeens} (13--18), with 146 native Arabic (L1) and 140 non-native Arabic (L2) speakers.

The L1 speakers represent four major Arabic dialectal backgrounds -- \textit{Egyptian}, \textit{Gulf}, \textit{Levantine}, and \textit{North African}, while the L2 speakers span diverse linguistic backgrounds across \textit{Asia}, \textit{Europe}, \textit{the Americas}, and \textit{Africa}. This design supports controlled analysis of age, L1/L2 background, and dialectal variation in speech and other NLP models. 
Figure~\ref{fig:corpus_overview} provides an overview of the corpus composition, while Figure~\ref{fig:dataset_characteristics} summarizes speaker distributions by age, gender, and nativity. Table~\ref{tab:avg_duration} reports the average utterance duration, and Table~\ref{tab:country_distribution} lists the top-10 countries by utterance count.


\begin{table}[t]
\centering
\caption{Average utterance duration (s) by age and nativity.}
\label{tab:avg_duration}
\footnotesize
\setlength{\tabcolsep}{7pt}
\renewcommand{\arraystretch}{0.9}

\begin{tabular}{lcc}
\toprule
& \textbf{7--12} & \textbf{13--18} \\
\midrule
Native     & 5.36 & 5.12 \\
Non-native & 8.45 & 6.67 \\
\bottomrule
\end{tabular}
\vspace{-0.25cm}
\end{table}








\paragraph*{Reading Material}
For recording, reading prompts are tailored to age, school grade, and Arabic proficiency. L1 \textsc{AraKids} use fully diacritized Grade 1--4 material, including stories, simple sentences, and isolated words for younger readers.
L1 \textsc{AraTeens}, on the other hand, use non-diacritized Grade 9--12 material with longer sentences and passages. L2 material is selected separately by age and Arabic proficiency, using simpler content for children and dedicated non-native Arabic material for adolescents.

\paragraph*{Data Collection and Metadata}
Speech was recorded on mobile devices through a dedicated collection platform, with guardian consent obtained for minors. Speaker metadata includes age, gender, nationality, mother tongue, L1/L2 status, years of learning Arabic, multilingual background, and education level, together with basic recording environment information. No personally identifiable information was retained. All recordings undergo quality control, and samples with incomplete or incorrect readings, excessive noise, clipping, or corrupted audio are removed.


\paragraph*{Data Splits}
We construct speaker-disjoint training, development, and test splits. The
training set includes L1 and L2 speakers from both \textsc{AraKids} and
\textsc{AraTeens}, while the development set is held out from training and
contains only L1 speakers.
We define two test conditions: \textsc{USUP}
(\textit{Unseen Speaker$\&$Unseen Prompt}), where both speakers and prompts are
unseen during training and development, and \textsc{USSP}
(\textit{Unseen Speaker$\&$Seen Prompt}), where unseen speakers read prompts that
also occur in training. \textsc{USUP} contains 2,500 utterances: 2,000 from
\textsc{AraKids} using 5--7 L1 speakers per region ($\approx$250 utterances),
and 500 from \textsc{AraTeens} using two speakers per region
($\approx$62 utterances). \textsc{USSP} uses the same speaker pool as
\textsc{USUP}, but with recordings of prompts seen during training. Detailed
statistics are given in Table~\ref{tab:dataset_stats}.

\paragraph*{Prompt--Verbatim Annotation}
For \textsc{USUP}, we have both the original reading prompt and a manually produced verbatim transcript of what the speaker actually said. Reading deviations are explicitly annotated as substitutions, insertions, and
deletions, while repetitions, stuttering, fillers, unclear speech, non-speech events, and speaker overlap are also marked. Dialectal and pronunciation-specific realizations are preserved, using extended Arabic characters when needed for sounds not represented in standard MSA orthography. These paired prompt--verbatim references support both ASR evaluation and fine-grained analysis of reading, pronunciation, and dialectal variation.

\begin{figure}[t]
    \centering


    \begin{subfigure}[t]{\columnwidth}
        \centering
        \includegraphics[width=\linewidth]{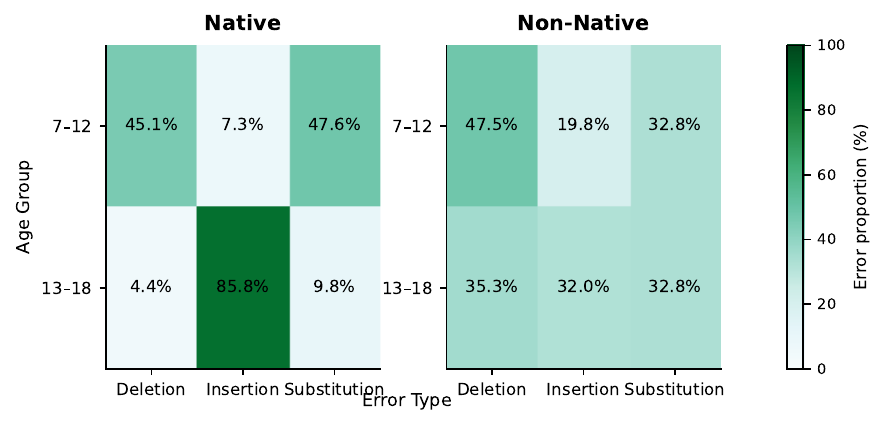}
        \label{fig:cer_age}
    \end{subfigure}

    \vspace{-0.1cm}
    \caption{Char-level reading deviations in the \textsc{USUP}
    set across age groups.}
    \label{fig:reading_errors}
    \vspace{-0.45cm}
\end{figure}

\paragraph*{Reading Deviation Analysis}

We then analyze the reading deviations in the manually transcribed \textsc{USUP} set by comparing each original prompt with its verbatim transcription. Character-level alignment is used to identify insertions, deletions, and substitutions, providing a fine-grained view of reading variation across L1/L2 speakers and age groups (see Figure~\ref{fig:reading_errors}). Reign-wise, L1 Egyptian speakers showed the highest deletion (62.7\%) and substitution (58.6\%) rates, whereas L2 speakers exhibited more balanced error patterns across regions.  
Our initial exploration of prompt--verbatim alignments reveals clear differences between L1 and L2 speakers. 

\noindent\textbf{Deletions} are the largest error category for both groups and are dominated by short-vowel and tanwīn drops. For L1 speakers, final short-vowel deletion is frequent and largely reflects pausal pronunciation and loss of grammatical markers (\textit{i'rāb}) rather than true reading errors. L2 speakers, however, show substantially more gemination (\textit{shadda}) drops than L1 speakers (120 vs.\ 21), consistent with difficulty in producing consonant length contrasts. They also omit long-vowel \textit{alif} more often (65 vs.\ 31), effectively reducing long vowels to short ones.

\noindent\textbf{Substitutions} are dominated by vowel changes. Consistent with the deletion patterns above, L1 speakers frequently reduce short-vowel endings to suk={u}n: 371/882 (42\%) of vowel changes are vowel$\rightarrow$suk{u}n, consistent with reduced final \textit{i\textsuperscript{`}r{a}b} realization. L2 speakers instead tend to substitute vowels toward the default fat'{h}a, with 184/534 (34\%) moving toward /a/ and {u}$\rightarrow${a} being the most common pattern (104, 19.5\%). The /u/$\rightarrow$/a/ pattern is also more prominent in L2 speakers (30\% vs. 14\%). Overall, L1 speakers \textbf{omit}, whereas L2 speakers \textbf{regularize} short vowels. Consonant substitutions, however, provide a clearer signal of speaker background: for L1 speakers, these patterns largely reflect dialectal variation, whereas for L2 speakers, they more often indicate phonological gaps in Arabic. Table~\ref{tab:substitution_patterns} summarizes these substitution patterns.

\noindent\textbf{Insertions} also differ across groups. For L2 speakers, \textit{fat={h}a} is the most frequent insertion, suggesting a default-vowel strategy when the intended vowel is uncertain. Insertions of \textit{alif} (\<أ>) and \textit{ya'} (\<ي>)  further suggest vowel lengthening, particularly among speakers from the Americas and Asia; the same pattern appears for L2-Teens, but less frequently. For L1-Teens, the high insertion rate is largely a diacritization artifact: inserted \textit{fat'{h}a}, \textit{kasra}, \textit{d{a}mma}, \textit{suk{u}n}, and \textit{shadda} occur across dialect regions because speakers restore diacritics absent from the undiacritized reference. These are scored as errors despite representing valid readings, while L1-Kids show fewer such insertions because their prompts are fully diacritized.

\begin{table}[t]
\centering
\caption{Top-10 countries by utterances.}
\label{tab:country_distribution}

\footnotesize
\setlength{\tabcolsep}{2.2pt}
\renewcommand{\arraystretch}{0.84}

\begin{tabular*}{\columnwidth}{
@{\extracolsep{\fill}}
rlr
rlr
@{}}
\toprule
\# & \textbf{Country} & \textbf{Utts. (\%) / Spk.} &
\# & \textbf{Country} & \textbf{Utts. (\%) / Spk.} \\
\midrule

1 & Egypt
& 14,315 (16.6\%) / 48
&
6 & Morocco
& 5,818 (6.7\%) / 17 \\

2 & India
& 8,081 (9.3\%) / 26
&
7 & Jordan
& 5,787 (6.7\%) / 21 \\

3 & France
& 7,448 (8.6\%) / 25
&
8 & Palestine
& 5,485 (6.3\%) / 16 \\

4 & U.S.
& 6,351 (7.3\%) / 20
&
9 & Pakistan
& 3,996 (4.6\%) / 14 \\

5 & Algeria
& 6,165 (7.1\%) / 19
&
10 & Iraq
& 3,462 (4.0\%) / 11 \\

\bottomrule
\end{tabular*}
\vspace{-0.5cm}
\end{table}











\begin{table}[t]
\centering
\caption{Frequent consonant substitutions with linguistic interpretation; Buckwalter transliteration and frequency are shown in parentheses. \<چ> uses extended symbol \texttt{c}.}
\label{tab:substitution_patterns}

\scriptsize
\setlength{\tabcolsep}{2.0pt}
\renewcommand{\arraystretch}{0.88}

\begin{tabularx}{\columnwidth}{
@{}>{\raggedright\arraybackslash}p{0.24\columnwidth}
   >{\raggedright\arraybackslash}p{0.31\columnwidth}
   X@{}}
\toprule
\textbf{Substitution} &
\textbf{Distribution} &
\textbf{Interpretation} \\
\midrule

\<ج>$\rightarrow$\<چ>\;{\tiny(\texttt{j$\rightarrow$c})}
& L1; Egypt (94);

L2; All regions (4)
& Dialectal variation \\

\rowcolor{gray!10}
\<ع>$\rightarrow$\<أ>\;{\tiny(\texttt{E$\rightarrow$>})}
& L2; All regions (63)
& Pharyngeal $\rightarrow$ glottal \\

\<ث>$\rightarrow$\<س>\;{\tiny(\texttt{v$\rightarrow$s})}
& L2; Africa/Asia (31); 

L1; Egypt/North Africa (16)
& Interdental $\rightarrow$ alveolar \\

\rowcolor{gray!10}
\<ث>$\rightarrow$\<ت>\;{\tiny(\texttt{v$\rightarrow$t})}
& L2 All reigns (23); L1; North Africa/Gulf (6)
& Interdental $\rightarrow$ alveolar \\

\<ذ>$\rightarrow$\<د>\;{\tiny(\texttt{*$\rightarrow$d})}
& L1; North Africa (3);

L2; All reigns (10)
& Dialectal/L2 variation \\

\rowcolor{gray!10}
\<ظ>$\rightarrow$\<ض>\;{\tiny(\texttt{Z$\rightarrow$D})}
& L1; North Africa (5)
& Emphatic merger \\

\<غ>$\rightarrow$\<چ>\;{\tiny(\texttt{g$\rightarrow$c})}
& L2; Africa (5)
& Uvular $\rightarrow$ velar \\

\rowcolor{gray!10}
\<ق>$\rightarrow$\<ك>\;{\tiny(\texttt{q$\rightarrow$k})}
& L2; Africa/America/Asia (5)
& Uvular $\rightarrow$ velar \\

\<ث>$\rightarrow$\<ت>\;{\tiny(\texttt{v$\rightarrow$t})}
& L1; North Africa (4)
& Interdental $\rightarrow$ alveolar \\

\rowcolor{gray!10}
\<ذ>$\rightarrow$\<ز>\;{\tiny(\texttt{*$\rightarrow$z})}
& L1; Egypt (3)
& Interdental $\rightarrow$ sibilant \\

\bottomrule
\end{tabularx}
\end{table}

\begin{table}[t]
\centering
\caption{Corpus statistics across splits. Per split reported utterances\ (speakers/hrs.).}
\label{tab:dataset_stats}

\footnotesize
\setlength{\tabcolsep}{1.8pt}
\renewcommand{\arraystretch}{0.98}

\begin{tabular*}{\columnwidth}{
@{\extracolsep{\fill}}lccccc@{}}
\toprule
\textbf{Split} &
\textbf{Kids-L1} & \textbf{Kids-L2} &
\textbf{Teens-L1} & \textbf{Teens-L2} &
\textbf{Total} \\
\midrule

Train &
\shortstack{12,572\\[-0.4ex]{\scriptsize(48/17.89)}} &
\shortstack{7,340\\[-0.4ex]{\scriptsize(24/16.41)}} &
\shortstack{19,705\\[-0.4ex]{\scriptsize(59/27.51)}} &
\shortstack{21,957\\[-0.4ex]{\scriptsize(68/41.13)}} &
\shortstack{61,574\\[-0.4ex]{\scriptsize(199/102.94)}} \\

Dev &
\shortstack{949\\[-0.4ex]{\scriptsize(3/1.58)}} &
\shortstack{--\\[-0.4ex]{\scriptsize(--/--)}} &
\shortstack{762\\[-0.4ex]{\scriptsize(2/1.18)}} &
\shortstack{--\\[-0.4ex]{\scriptsize(--/--)}} &
\shortstack{1,711\\[-0.4ex]{\scriptsize(5/2.75)}} \\

\textsc{USUP} &
\shortstack{1,000\\[-0.4ex]{\scriptsize(26/1.57)}} &
\shortstack{1,000\\[-0.4ex]{\scriptsize(40/2.43)}} &
\shortstack{250\\[-0.4ex]{\scriptsize(8/0.40)}} &
\shortstack{250\\[-0.4ex]{\scriptsize(8/0.38)}} &
\shortstack{2,500\\[-0.4ex]{\scriptsize(82/4.78)}} \\

\textsc{USSP} &
\shortstack{7,242\\[-0.4ex]{\scriptsize(26/11.40)}} &
\shortstack{10,061\\[-0.4ex]{\scriptsize(40/24.36)}} &
\shortstack{1,839\\[-0.4ex]{\scriptsize(8/2.97)}} &
\shortstack{1,544\\[-0.4ex]{\scriptsize(8/2.52)}} &
\shortstack{20,686\\[-0.4ex]{\scriptsize(82/41.25)}} \\

\midrule

\textbf{All} &
\shortstack{21,763\\[-0.4ex]{\scriptsize(77/32.43)}} &
\shortstack{18,401\\[-0.4ex]{\scriptsize(64/43.20)}} &
\shortstack{22,556\\[-0.4ex]{\scriptsize(69/32.07)}} &
\shortstack{23,751\\[-0.4ex]{\scriptsize(76/44.03)}} &
\shortstack{\textbf{86,471}\\[-0.4ex]{\scriptsize\textbf{(286/151.72)}}} \\

\bottomrule
\end{tabular*}

\vspace{-0.6cm}

\end{table}

\begin{table*}[!ht]
\centering
\caption{WER (\%) on \aranextgenvoices{}.
\textsc{USUP} reports
\textcolor{verbatimblue}{$\mathrm{WER}_{V}$}\,/\, $\mathrm{WER}_{P}$,
where $V$ and $P$ denote the human-corrected verbatim transcription and
reading prompt and \textsc{USSP} reports $\mathrm{WER}_{P}$.
ZS: zero-shot; AK, AT, and JT: fine-tuning on
\textsc{AraKids}, \textsc{AraTeens}, and joint data.
Best results are in \textbf{bold}.}
\label{tab:main_results}

\fontsize{7.4}{8.0}\selectfont
\setlength{\tabcolsep}{1.0pt}
\renewcommand{\arraystretch}{0.82}

\makebox[\textwidth][c]{%
\begin{minipage}[t]{0.49\textwidth}
\centering

\begin{tabular*}{\linewidth}{
@{\extracolsep{\fill}}llcccc@{}}
\toprule
\textbf{Model} & \textbf{FT} &
\textbf{Kids-L1} & \textbf{Kids-L2} &
\textbf{Teens-L1} & \textbf{Teens-L2} \\
\midrule

\multicolumn{6}{c}{
\textbf{Useen-Speaker$\&$Unseen-Prompt \textsc{USUP}}\;
(\textcolor{verbatimblue}{$\mathrm{WER}_{V}$}\,/\, $\mathrm{WER}_{P}$)}
\\[-0.3ex]
\midrule

\multirow{4}{*}{\shortstack{Aura-Off.\\{\scriptsize 342M}}}
& ZS
& \textcolor{verbatimblue}{10.71}\,/\,8.25
& \textcolor{verbatimblue}{39.60}\,/\,33.90
& \textcolor{verbatimblue}{11.64}\,/\,8.38
& \textcolor{verbatimblue}{36.79}\,/\,28.92 \\

& AK
& \textcolor{verbatimblue}{\textbf{5.57}}\,/\,\textbf{2.98}
& \textcolor{verbatimblue}{21.16}\,/\,13.39
& \textcolor{verbatimblue}{10.77}\,/\,7.28
& \textcolor{verbatimblue}{26.31}\,/\,16.37 \\

& AT
& \textcolor{verbatimblue}{9.49}\,/\,7.35
& \textcolor{verbatimblue}{31.72}\,/\,25.79
& \textcolor{verbatimblue}{12.75}\,/\,9.20
& \textcolor{verbatimblue}{23.41}\,/\,13.34 \\

& JT
& \textcolor{verbatimblue}{8.74}\,/\,6.32
& \textcolor{verbatimblue}{24.94}\,/\,17.53
& \textcolor{verbatimblue}{13.50}\,/\,10.19
& \textcolor{verbatimblue}{22.41}\,/\,11.55 \\

\midrule

\multirow{4}{*}{\shortstack{Cohere-Ar\\{\scriptsize 2B}}}
& ZS
& \textcolor{verbatimblue}{6.60}\,/\,4.60
& \textcolor{verbatimblue}{42.88}\,/\,37.70
& \textcolor{verbatimblue}{\textbf{9.90}}\,/\,\textbf{6.69}
& \textcolor{verbatimblue}{39.91}\,/\,32.62 \\

& AK
& \textcolor{verbatimblue}{6.65}\,/\,3.91
& \textcolor{verbatimblue}{20.57}\,/\,12.94
& \textcolor{verbatimblue}{12.22}\,/\,8.91
& \textcolor{verbatimblue}{25.86}\,/\,16.70 \\

& AT
& \textcolor{verbatimblue}{7.56}\,/\,5.13
& \textcolor{verbatimblue}{29.33}\,/\,22.69
& \textcolor{verbatimblue}{10.54}\,/\,6.93
& \textcolor{verbatimblue}{22.63}\,/\,14.91 \\

& JT
& \textcolor{verbatimblue}{6.55}\,/\,3.98
& \textcolor{verbatimblue}{\textbf{20.35}}\,/\,\textbf{12.92}
& \textcolor{verbatimblue}{10.13}\,/\,\textbf{6.69}
& \textcolor{verbatimblue}{\textbf{20.62}}\,/\,\textbf{11.43} \\

\midrule

\multicolumn{6}{c}{
\textbf{Useen-Speaker$\&$Seen-Prompt \textsc{USSP}}\; ($\mathrm{WER}_{P}$)}
\\[-0.3ex]
\midrule

\multirow{4}{*}{\shortstack{Aura-Off.\\{\scriptsize 342M}}}
& ZS & 7.24 & 34.03 & 9.89 & 27.71 \\
& AK & \textbf{2.10} & \textbf{9.76} & 9.18 & 16.43 \\
& AT & 8.36 & 25.77 & \textbf{3.49} & 2.92 \\
& JT & 2.36 & 12.16 & 3.55 & 3.25 \\

\midrule

\multirow{4}{*}{\shortstack{Cohere-Ar\\{\scriptsize 2B}}}
& ZS & 4.37 & 39.97 & 7.65 & 28.89 \\
& AK & 2.66 & 9.79 & 10.25 & 19.47 \\
& AT & 5.19 & 21.55 & 4.60 & 3.28 \\
& JT & 2.44 & 9.87 & 4.64 & 2.90 \\

\bottomrule
\end{tabular*}
\end{minipage}%
\hspace{0.02\textwidth}%
%
\begin{minipage}[t]{0.49\textwidth}
\centering

\begin{tabular*}{\linewidth}{
@{\extracolsep{\fill}}llcccc@{}}
\toprule
\textbf{Model} & \textbf{FT} &
\textbf{Kids-L1} & \textbf{Kids-L2} &
\textbf{Teens-L1} & \textbf{Teens-L2} \\
\midrule

\multicolumn{6}{c}{
\textbf{Useen-Speaker$\&$Unseen-Prompt \textsc{USUP}}\;
(\textcolor{verbatimblue}{$\mathrm{WER}_{V}$}\,/\, $\mathrm{WER}_{P}$)}
\\[-0.3ex]
\midrule

\multirow{4}{*}{\shortstack{Qwen-O-Ar\\{\scriptsize 7B}}}
& ZS
& \textcolor{verbatimblue}{9.57}\,/\,7.17
& \textcolor{verbatimblue}{37.83}\,/\,32.49
& \textcolor{verbatimblue}{14.44}\,/\,11.58
& \textcolor{verbatimblue}{31.33}\,/\,23.43 \\

& AK
& \textcolor{verbatimblue}{10.53}\,/\,8.33
& \textcolor{verbatimblue}{27.99}\,/\,21.90
& \textcolor{verbatimblue}{23.28}\,/\,20.95
& \textcolor{verbatimblue}{29.54}\,/\,21.52 \\

& AT
& \textcolor{verbatimblue}{14.85}\,/\,12.65
& \textcolor{verbatimblue}{38.74}\,/\,34.02
& \textcolor{verbatimblue}{18.10}\,/\,14.84
& \textcolor{verbatimblue}{28.76}\,/\,21.64 \\

& JT
& \textcolor{verbatimblue}{11.15}\,/\,8.83
& \textcolor{verbatimblue}{25.37}\,/\,18.99
& \textcolor{verbatimblue}{15.95}\,/\,12.81
& \textcolor{verbatimblue}{26.09}\,/\,17.83 \\

\midrule

\multirow{4}{*}{\shortstack{Whisper-v3\\{\scriptsize 1.55B}}}
& ZS
& \textcolor{verbatimblue}{11.81}\,/\,9.76
& \textcolor{verbatimblue}{47.15}\,/\,43.00
& \textcolor{verbatimblue}{14.32}\,/\,12.22
& \textcolor{verbatimblue}{38.35}\,/\,32.29 \\

& AK
& \textcolor{verbatimblue}{10.31}\,/\,7.99
& \textcolor{verbatimblue}{26.61}\,/\,20.06
& \textcolor{verbatimblue}{19.97}\,/\,16.88
& \textcolor{verbatimblue}{30.55}\,/\,22.42 \\

& AT
& \textcolor{verbatimblue}{13.29}\,/\,11.21
& \textcolor{verbatimblue}{36.28}\,/\,31.27
& \textcolor{verbatimblue}{13.56}\,/\,10.48
& \textcolor{verbatimblue}{24.08}\,/\,15.02 \\

& JT
& \textcolor{verbatimblue}{9.98}\,/\,7.78
& \textcolor{verbatimblue}{26.86}\,/\,20.55
& \textcolor{verbatimblue}{13.39}\,/\,9.95
& \textcolor{verbatimblue}{25.08}\,/\,16.37 \\

\midrule

\multicolumn{6}{c}{
\textbf{Useen-Speaker$\&$Seen-Prompt \textsc{USSP}}\; ($\mathrm{WER}_{P}$)}
\\[-0.3ex]
\midrule

\multirow{4}{*}{\shortstack{Qwen-O-Ar\\{\scriptsize 7B}}}
& ZS & 7.61 & 34.21 & 15.07 & 26.80 \\
& AK & 2.24 & 12.57 & 23.28 & 31.86 \\
& AT & 13.31 & 34.21 & 3.81 & 1.48 \\
& JT & 2.38 & 11.63 & 5.22 & \textbf{1.40} \\

\midrule

\multirow{4}{*}{\shortstack{Whisper-v3\\{\scriptsize 1.55B}}}
& ZS & 10.02 & 40.16 & 15.21 & 40.71 \\
& AK & 4.42 & 14.37 & 18.22 & 26.38 \\
& AT & 10.82 & 29.14 & 6.59 & 2.62 \\
& JT & 4.28 & 14.58 & 6.73 & 2.86 \\

\bottomrule
\end{tabular*}
\vspace{-0.8cm}
\end{minipage}%
}
\end{table*}

\section{Experiment Setup}
\label{sec:experiments}

\paragraph*{Models}
We evaluate four recent Arabic and multilingual speech models: \textsc{Aura-STT} (347M),\footnote{\url{https://www.api.fanar.qa/}} \textsc{Cohere Transcribe Arabic} (2B),\footnote{\url{https://huggingface.co/CohereLabs/cohere-transcribe-03-2026}} \textsc{Qwen2.5-Omni-Arabic} (7B) \cite{bhatti2026multitaskinstructiontuningdata}, and \textsc{Whisper Large-v3} (1.55B) \cite{radford2022robustspeechrecognitionlargescale}. The set includes dedicated Arabic ASR models and general-purpose omni models, allowing us to compare Arabic specialization along with age and nativity.

\paragraph*{Adaptation Strategy}
We first evaluate each model in a zero-shot (ZS) setting without adaptation. We then fine-tune using three configurations: \textsc{AraKids} only (AK), \textsc{AraTeens} only (AT), and the joint training data (JT). The same
training splits are used across models, with speakers kept disjoint between training and evaluation. 

\subsection{Evaluation Measure}
\label{sec:eval_protocol}

We evaluate all systems on \textsc{USUP} and \textsc{USSP} using WER and CER. The same Arabic normalization is applied to the reading prompt ($P$), verbatim transcription ($V$), and ASR hypothesis. Verbatim markup for non-speech events, fillers, overlap, unintelligible speech, speaker labels, and control tokens is removed, while repetitions are preserved. We then remove diacritics, tatweel, punctuation, and normalize Alef variants (\<أ>, \<إ>, \<آ>, \<ٱ>) to \<ا> (A). Dialectal letters such as\footnote{Special letters are not found in Buckwalter transliteration.} \<چ>, \<ڟ>, \<ڤ>, \<گ>, and \<پ> are kept, as are the distinctions \<ة>/\<ه> (p/h) and
\<ى>/\<ي> (Y/y).

\section{Results and Analysis}
\label{sec:results}
Table~\ref{tab:main_results} compares ASR performance across age, L1/L2 background, and fine-tuning settings.


\paragraph*{Age and L1/L2 Effects}
Our results show that L2 speech is consistently more difficult than L1 speech across all models. In the ZS setting, L2 children also have higher WER than L2 adolescents. Fine-tuning substantially reduces this gap, particularly for L2 speech. On \textsc{USSP}, age-matched fine-tuning generally favors the target group such that AK performs improves for \textsc{AraKids}, while AT favors \textsc{AraTeens}. On the more difficult \textsc{USUP} test set, joint fine-tuning is often competitive or better, suggesting a stronger balance across the full 7--18 age range.

\begin{table}[t]
\centering
\caption{Representative dialectal substitutions seen in ASR; Buckwalter
transliteration is shown in parentheses.}
\label{tab:dialectal_patterns}

\scriptsize
\setlength{\tabcolsep}{2pt}
\renewcommand{\arraystretch}{0.82}

\begin{tabular*}{0.90\columnwidth}{
@{\extracolsep{\fill}}lll@{}}
\toprule
\textbf{Pattern} & \textbf{Group} & \textbf{Type} \\
\midrule

\<ع>$\rightarrow$\<ا>\;{\tiny(\texttt{E$\rightarrow$A})}
& L2
& Pharyngeal \\

\rowcolor{gray!10}
\<ق>$\leftrightarrow$\<ك>,\;
\<ق>$\rightarrow$\<غ>\;{\tiny(\texttt{q$\leftrightarrow$k, q$\rightarrow$g})}
& L2
& Uvular \\

\<غ>$\rightarrow$\<ج>\;{\tiny(\texttt{g$\rightarrow$j})}
& Gulf
& Dialectal \\

\rowcolor{gray!10}
\<ط>$\rightarrow$\<ت>\;{\tiny(\texttt{T$\rightarrow$t})}
& Gulf
& De-emphasis \\

\<ض>$\rightarrow$\<ظ>\ /\ \<ز>\ /\ \<د>\;{\tiny(\texttt{D$\rightarrow$Z/z/d})}
& North Africa
& Emphatic \\

\rowcolor{gray!10}
\<ث>$\leftrightarrow$\<س> /\<ت>\;{\tiny(\texttt{v$\leftrightarrow$s/t})}
& L1
& Interdental \\

\<ة>$\rightarrow$\<ت>\;{\tiny(\texttt{p$\rightarrow$t})}
& L1/L2
& Connected speech \\

\bottomrule
\end{tabular*}
\end{table}

\paragraph*{Prompt vs.\ Verbatim Speech}
For every \textsc{USUP} setting, $\mathrm{WER}_{P}$ is lower than $\mathrm{WER}_{V}$, with a larger difference for L2 speech. This shows that ASR outputs are often closer to the intended reading prompt than to the speaker's verbatim transcription. This is consistent with partial normalization of reading deviations. We analyze the prompt ($P$), verbatim transcription ($V$), and ASR hypothesis ($H$) jointly to distinguish preserved deviations ($H\!\approx\!V$) from normalization toward the prompt ($H\!\approx\!P$).
\paragraph*{Pronunciation and Error Patterns}
Manual alignment reveals systematic age-, L2-, and dialect-related variation. L2 children, for example, show more frequent deletions and insertions involving \<ا> and \<ل>, while several consonant
substitutions reflect regional or L2 pronunciation patterns (Table~\ref{tab:dialectal_patterns}). We also observe model-related orthographic errors, including frequent \<ا> insertions and hamza-seat
variations such as \<ئ>$\rightarrow$\<ء> (\texttt{\}}$\rightarrow$') and
\<ء>$\rightarrow$\<ؤ> ('$\rightarrow$\texttt{\&}). These should be distinguished from genuine pronunciation differences when evaluating young-speaker ASR.
\paragraph*{Key Findings}
Overall, L2 speech remains substantially more challenging than L1 speech, with majority of the errors coming from younger L2 speakers. Age-specific fine-tuning improve the matched age group performance, while joint fine-tuning shows a stronger balance across age-nativity groups. Moreover, $\mathrm{WER}_{P}$ is consistently lower than $\mathrm{WER}_{V}$, particularly for L2 speakers, suggesting that ASR outputs often move toward the intended prompt rather than fully preserving the dialectal information. Error patterns further show systematic age-, L2-, and dialect-related pronunciation differences alongside model-specific orthographic and decoding errors.

\section{Conclusions}
\label{sec:conclusions}

We introduced \aranextgenvoices{}, a diverse Arabic speech corpus and, to our knowledge, the first to jointly cover child and adolescent speech from ages 7--18 across both L1 and L2 speakers. Follwoing, we benchmarked the datasets with state-of-the-art ASRs in zero-, and fine-tuned setting.
Our results show substantial performance gaps for L2 speech, particularly among younger speakers, while age-specific and joint fine-tuning significantly improve recognition. Analysis of prompt--verbatim differences further reveals systematic age-, dialect-, and L2-related reading variation, as well as a tendency of ASR systems to normalize some deviations toward the intended prompt. We hope this work supports more robust and inclusive Arabic ASR for young speakers.

\paragraph*{Limitation and Distribution:} A limitation of the \aranextgenvoices{} is that it focuses on read speech and is not always uniformly balanced across age, regional, and L1/L2 groups. Moreover, as the recordings involve minors and include demographic and linguistic metadata, the raw audio and full speaker-level metadata will not be publicly released. We will instead release de-identified aggregate statistics, evaluation protocols, and resulting ASR models, subject to the applicable consent and institutional requirements.



\newpage

\bibliographystyle{IEEEbib}
\bibliography{refs}




\end{document}